\documentclass[compsoc,onecolumn]{IEEEtran}
\pdfoutput=1
\usepackage{graphicx}
\usepackage[caption=false,font=footnotesize,labelfont=sf,textfont=sf]{subfig}
\usepackage[font={normalsize,sf},labelfont=bf,labelsep=period]{caption}
\usepackage{tabulary}
\usepackage{url}
\usepackage{natbib}

\newcommand\EatDot[1]{}

\begin{document}

\title{Mining Spatio-temporal Data on Industrialization from Historical Registries}

\author{
  \normalsize
  \IEEEauthorblockN{
    David Berenbaum\IEEEauthorrefmark{1},
    Dwyer Deighan\IEEEauthorrefmark{1},
    Thomas Marlow\IEEEauthorrefmark{2}, 
    Ashley Lee\IEEEauthorrefmark{1},
    Scott Frickel\IEEEauthorrefmark{2} and
    Mark Howison\IEEEauthorrefmark{1}
  }\\[2ex]
  \IEEEauthorblockA{
    \begin{tabular}{cc}
      \IEEEauthorrefmark{1} Data Science Practice
    &
      \IEEEauthorrefmark{2} Institute at Brown for Environment and Society
    \\
      Computing \& Information Services
    &
      Brown University
    \\
      Brown University
    &
      80 Waterman Street
    \\
      3 Davol Square
    &
      Providence, RI 02912
    \\
      Providence, RI 02912
    \end{tabular}
  }\\[2ex]
  \IEEEauthorblockA{
    Email: \{david\_berenbaum, dwyer\_deighan, thomas\_marlow, ashley\_lee, scott\_frickel, mhowison\}@brown.edu
  }
}

\maketitle

\begin{abstract}
Despite the growing availability of big data in many fields, historical data on socioevironmental phenomena are often not available due to a lack of automated and scalable approaches for collecting, digitizing, and assembling them. We have developed a data-mining method for extracting tabulated, geocoded data from printed directories.
While scanning and optical character recognition (OCR) can digitize printed text, these methods alone do not capture the structure of the underlying data.
Our pipeline integrates both page layout
analysis and OCR to extract tabular, geocoded data from structured text.
We demonstrate the utility of this method by applying it to
scanned manufacturing registries from Rhode Island that record 41 years of
industrial land use.  The resulting spatio-temporal data can be used for
socioenvironmental analyses of industrialization at a resolution that was not
previously possible. In particular, we find strong evidence for the dispersion
of manufacturing from the urban core of Providence, the state's capital, along
the Interstate 95 corridor to the north and south.
\end{abstract}

\noindent

\section{Introduction}

In most states in the U.S., detailed registries of manufacturers are compiled
annually, dating back to the 1950s. These printed historical registries are a
rich source of data on the location, size, and type of industrial activity over
time. However, mining that data is not straight-forward using existing OCR
tools, because of the registries' structure.  To address this, we developed
\texttt{georeg}, a
pipeline for extracting addresses and other business information from
historical registries, and tested it with images we scanned from Rhode Island
manufacturing registries spanning the 1950s through the 1990s.  In these
scanned images, \texttt{georeg} identifies each heading and the ordering of
manufacturer listings, so that we can extract the name, address, business type,
and number of employees as tabular data, which is then geocoded to provide
latitude and longitude.

With this spatio-temporal data, we are able to perform more detailed
socio-environmental analyses of changes in industrialization and locations of
potentially hazardous manufacturing sites than was previously possible. Among
other findings, the data show a dispersion of industrial sites from Providence,
the state's capital and most populous city, to adjoining cities along the
Interstate 95 corridor.

\section{Background}

While many robust methods exist for digitizing historical documents, there has
been less focus on modeling the complex structure in printed texts like
directories, which use combinations of headings, font styles, and column and
block layouts to organize and present repeated listings of structured data.
The two primary components of digitizing printed documents are page layout
analysis and optical character recognition (OCR). Page layout analysis provides
information about the position and arrangement of blocks of text within a page.
OCR converts pixelated images into characters and words.

While OCR makes it possible to extract all text from scanned images of
directories, it fails to automatically detect complex structure, even when
using more advanced OCR methods that incorporate page layout analysis, such as
those in the Tesseract package \citep{Smith_2009}. Tesseract includes an OCR
mode that detects tab-stops to extract text that is consistent with the
document's column layout. However, this approach does not segment individual
text blocks within the columns.

Although there is an extensive literature on page layout analysis (for a
review, see \citet{Tang_Lee_Suen_1996} or \citet{Mao_Rosenfeld_Kanungo_2003}),
existing approaches to segmenting documents are largely based on physical
features such as the relative position and size of lines or blocks of text. In
general, physical segmentation does not capture the hierarchy or logical
structure of a complex document.  The existing approach that is closest to ours
is a method created by \citet{Ma_Doermann_2003} that segments
documents with repeated entries of similar structure, such as phone books and
dictionaries.  In this method, both physical features (e.g., line structure)
and logical features (e.g., word patterns) are used to train a segmentation
model, starting from sample documents which have already been correctly
segmented into repeated entries.  However, the segmentation provided by this
method is limited in that it does not distinguish between different kinds of
entries (e.g. headings vs. listings) and does not capture the structure of data
elements within an entry (e.g. name vs. address vs. phone number in a phone
book entry).

\section{Methods}

\subsection{Scanning}

We scanned printed registries through a combination of book scanners and
photography. Manufacturing registries for 16 individual years between 1953 and
1994 were accessed at the Brown University Library and the Rhode Island
Historical Society. We used the open-source software ScanTailor
(\url{http://scantailor.org}) to trim edges, straighten, de-warp, and convert
the images to grayscale.

\subsection{Contour merging}

We developed a contour merging algorithm to partition a page image into the
areas that correspond to individual records.  Contours are
the outermost boundaries of a set of contiguous pixels of the same intensity
\citep{suzuki_topological_1985} and we initially identify them using OpenCV
\citep{opencv_library}.
However, these contours frequently outline individual characters in the
text. We use thresholding, erosion, and dilation to merge nearby contours into
contiguous areas of the image that cover all characters in a
record (Figure~\ref{fig:closed}).

After thresholding and inverting the grayscale image to white text on a black
background, we repeatedly perform a close operation, which is an erosion
followed by a dilation. The close operations eliminate small gaps of
black-space between characters and lines of text. Then, we perform fewer
iterations of an open operation, which is a dilation followed by an erosion.
The open operations de-noise the contours and smooth them, removing small
artifacts of white pixels.
For the erosion and dilation operation's kernel shape, we use
a rectangle with dimensions that the user can configure. Formally, binary
erosion of image $I$ by kernel $K$ is
defined as: \[I \ominus K = \{p | K_p \subseteq I\},\] where $K_p$ is the
kernel for pixel $p$. That is, each pixel $p$ remains white only if every
other pixel in its kernel is also white. Similarly, binary dilation is defined
as: \[I \oplus  K = \bigcup _{p \in I} K_p.\]

In some cases, there is no additional space between the lines of text for each
manufacturer record. In these registry formats, a new record is denoted by
indenting the first line. When we merge the contours of these pages, a single
merged contour may encompass several records or an entire column of text. In
this case, we use the indentation to identify individual records and split the
contour at each indentation.  We identify each indent by following the contour
and noting where the x-coordinate stretches beyond a given threshold distance
from the left bound of the contour.

\subsection{Identifying manufacturer records}

\begin{figure*}[p]
\centering
\subfloat[Dilation/Erosion\label{fig:closed}]{\includegraphics[width=0.24\linewidth]{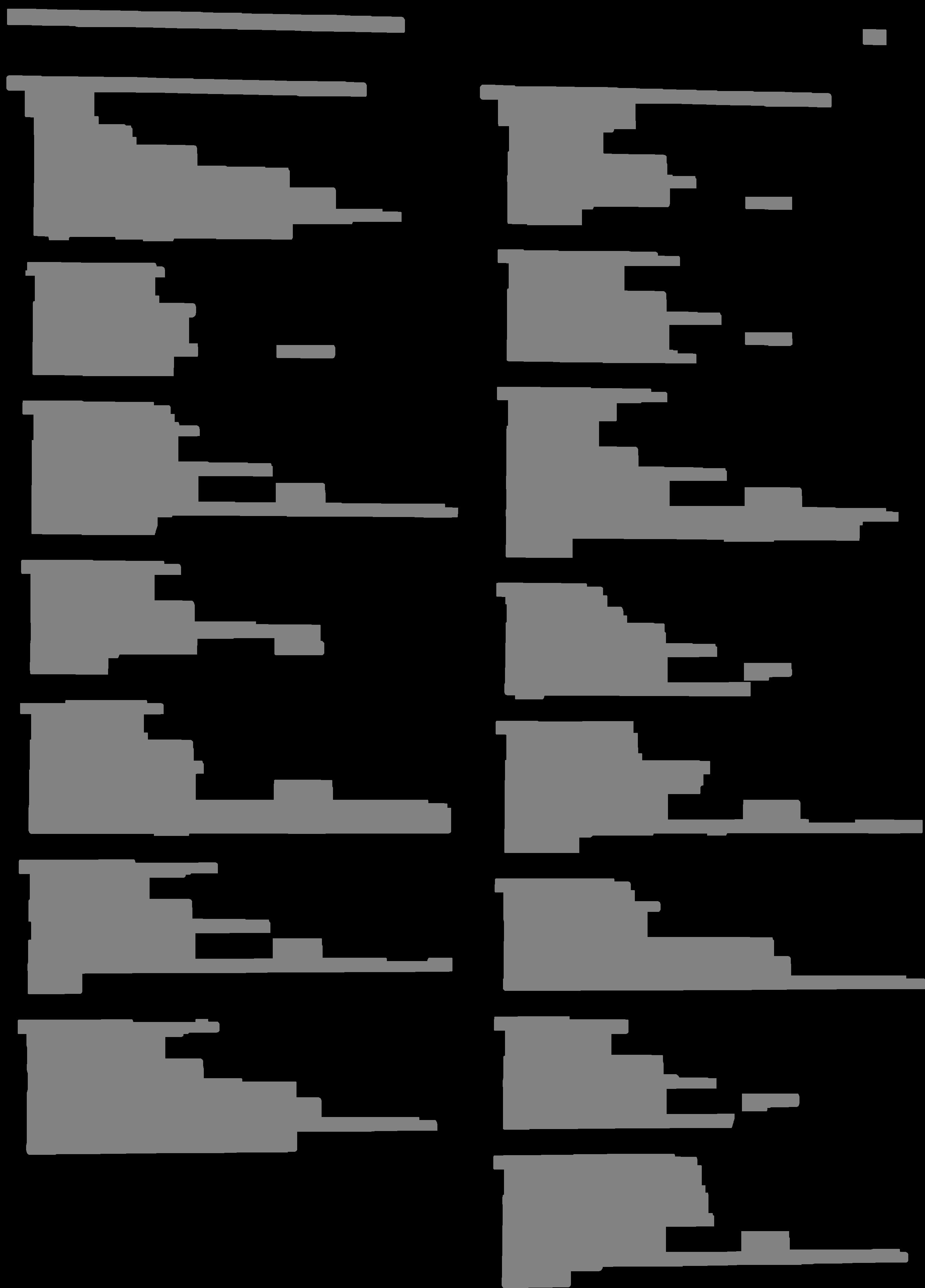}}
\hfill
\subfloat[Center Clustering\label{fig:clusters}]{\includegraphics[width=0.24\linewidth]{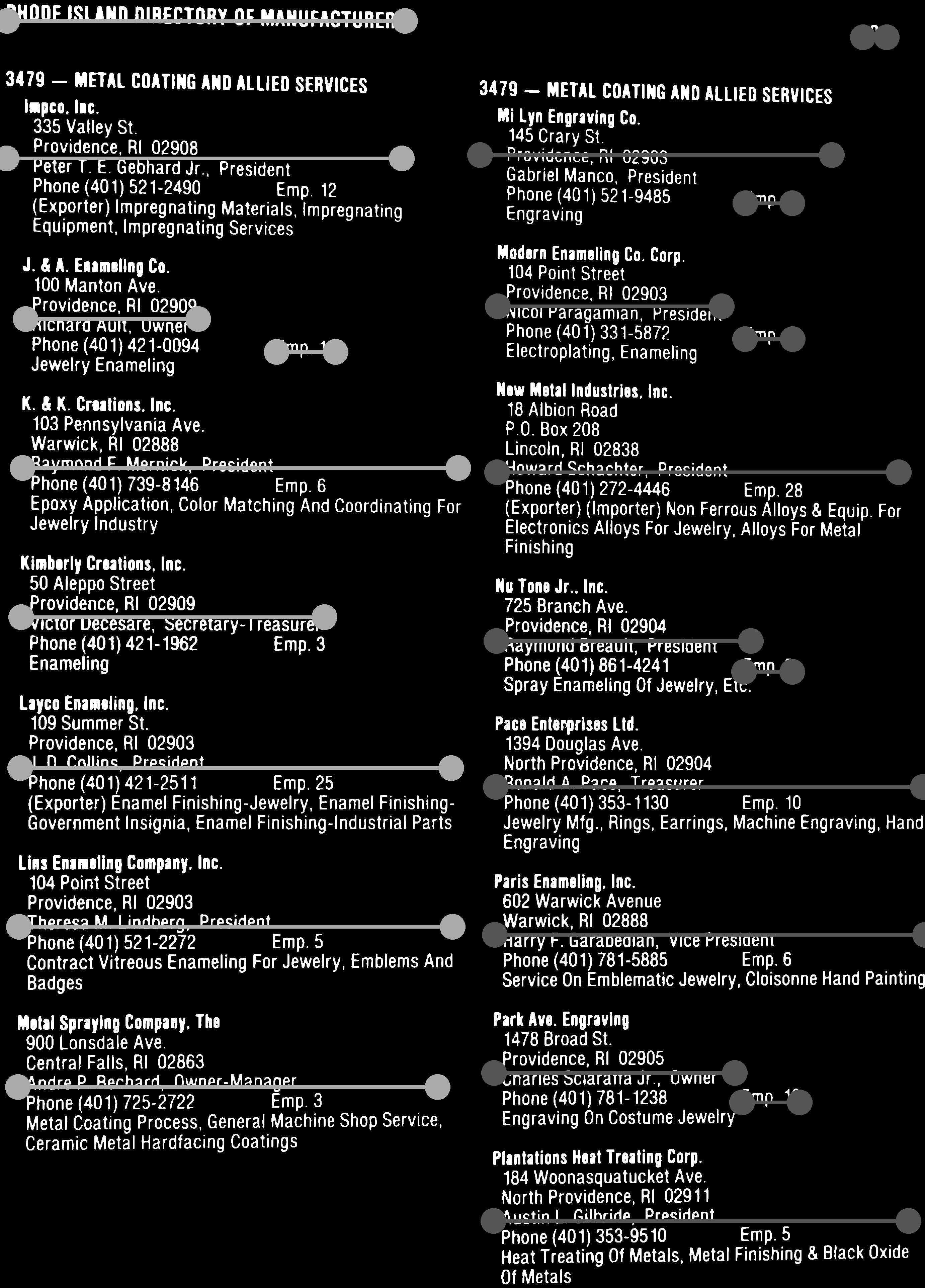}}
\hfill
\subfloat[Column Positions\label{fig:column_lines}]{\includegraphics[width=0.24\linewidth]{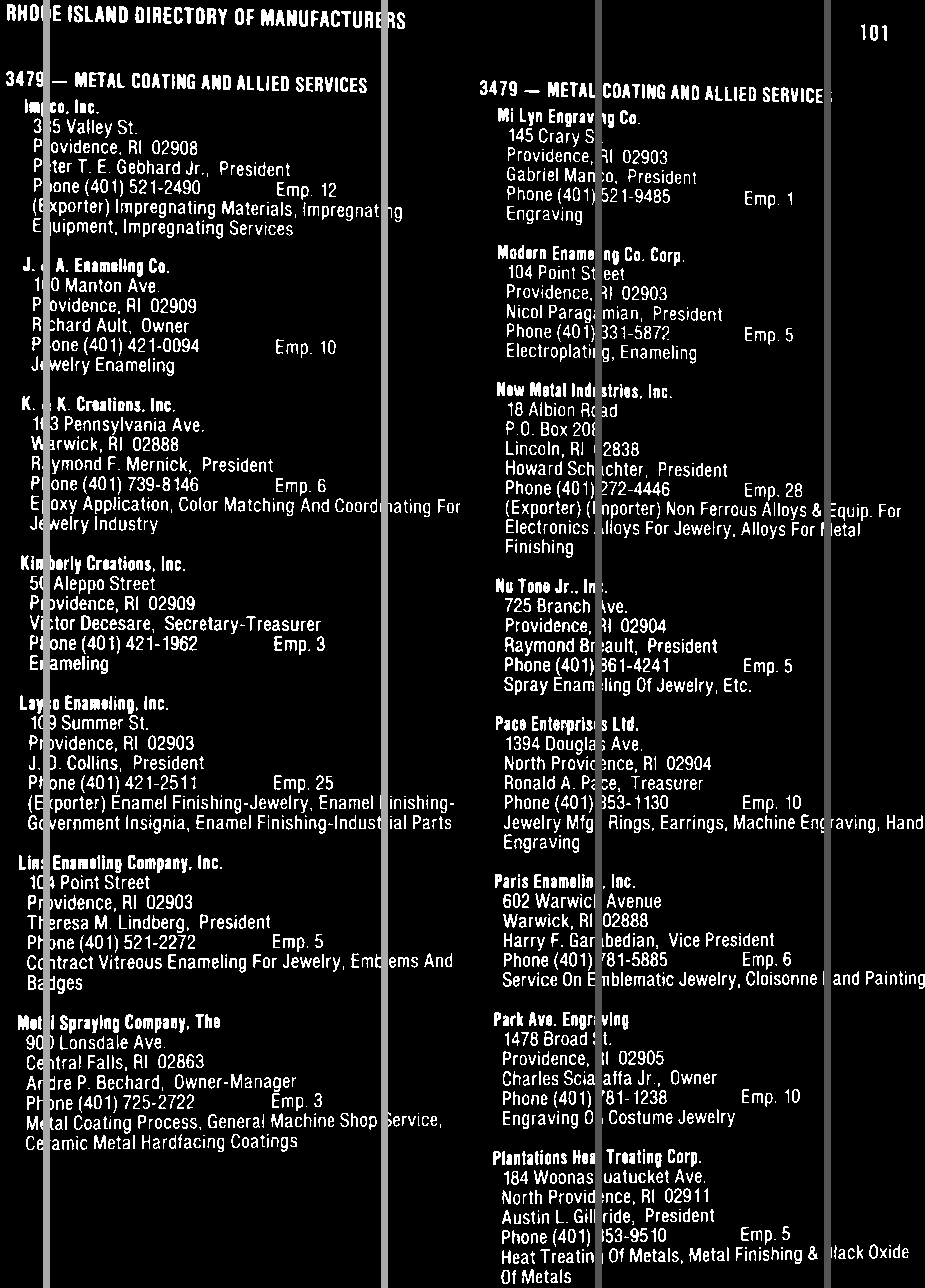}}
\hfill
\subfloat[Headings/Records for OCR\label{fig:contoured}]{\includegraphics[width=0.24\linewidth]{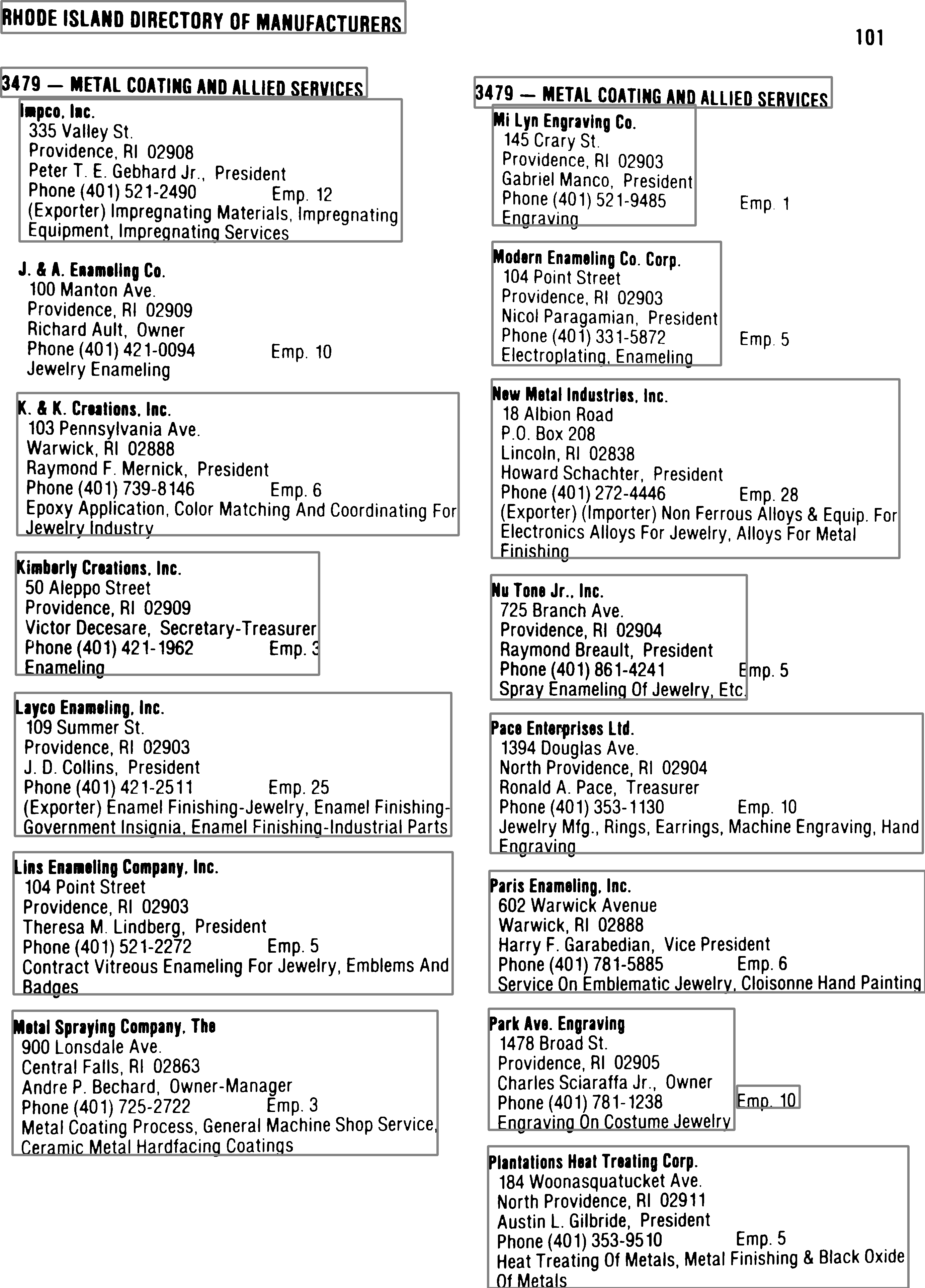}}
\caption{A contour merging and column clustering approach to identifying headings and records in the registry pages.}
\end{figure*}

After merging the contours, some correspond to manufacturer records, but
others are image artifacts or correspond to extraneous information, like
page numbers. The manufacturer records are bounded within the columns of the
page, so we only consider a contour as a candidate for a manufacturer record
if it is aligned with a page column.  We use k-means clustering
\citep{lloyd_least_1982} to group the contours by their left and right
bounds, setting the number of clusters equal to the number of columns in the
page (Figure~\ref{fig:clusters}). Column-aligned listings have left and right
bounds close to the column boundaries and generally outnumber the extraneous
contours, so the clusters converge around the left and right bounds of the
columns (Figure~\ref{fig:column_lines}). To filter out extraneous contours, we
eliminate those which are greater than a given number of standard deviations
away from the cluster centroids.

\subsection{Parsing information from records}

For each potential manufacturer record, we perform OCR with
Tesseract~\citep{Smith_2007} on the sub-image defined by the bounding box
of the contour (Figure~\ref{fig:contoured}).  We
use regular expressions to extract the business name, type, address, number of
employees, and other standardized information included for each record. If the
text is more consistent with a heading (for example, it contains only a city
name), we retain that information and include it with each subsequent
manufacturer record under the heading.

Some headings are centered across the page, in which case they are not bounded
within the page columns and are eliminated from potential manufacturer
records.  If such headings exist and contain meaningful information, we OCR
all headings centered across the page and note their vertical positions. For
each manufacturer record, we include the information from the closest heading 
above it (or the last heading on a previous page if there is no heading above 
it).

\subsection{Geocoding}

For each manufacturer record in which an address was identified, we geocode
the address to obtain a latitude, longitude, and confidence score. To improve
geocoding success, we match the cities to a list of correctly spelled cities
within the state. If the match ratio (based on Levenshtein distance) for the
best city match is above a given threshold, we replace the city name found by
OCR with the matched city name. We geocode each address using a local ArcGIS
server at Brown University, but any geocoding service supported by the
\texttt{geopy} package could be substituted.
\section{Results}

\subsection{Accuracy}

\begin{figure*}[p]
\centering
\includegraphics[width=\linewidth]{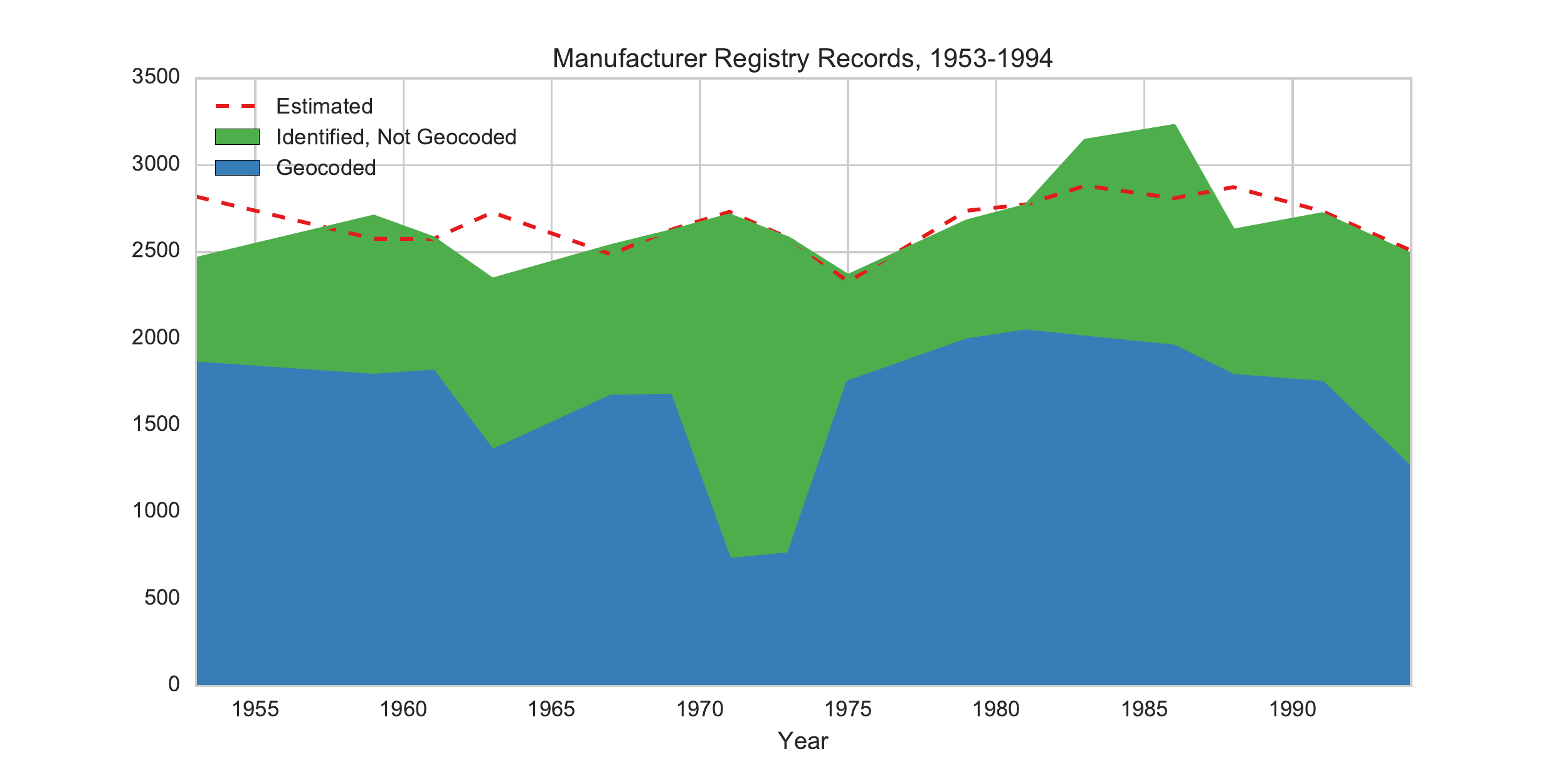}
\caption{The number of manufacturers identified and geocoded in each year's registry, compared to the estimated number of manufacturers.}
\label{fig:georeg_accuracy}
\end{figure*}

To evaluate the accuracy of \texttt{georeg},
we compared the number of
identified manufacturer records with an estimate of the actual content of the
registries. For the estimate, we manually counted the number of records on ten
pages of each year's directory and then multiplied the average number of
records per page by the total page count.
Overall, \texttt{georeg} identified 99\% of the estimated number of records
across all pages tested (Figure~\ref{fig:georeg_accuracy}). Individual years
differ in how closely they approximate the estimated number of records, but
each year's results are within 15\% of the estimate.

We also measured the number of records in which the address was geocoded with
a confidence score of at least 75\%. \texttt{georeg} successfully geocoded
61\% of the estimated number of records, but the results differed dramatically
by year. In 12 of the 16 years, between 62\% and 75\% of the estimated number
of records were successfully geocoded. However, in 1971, only 26\% were
successfully geocoded, and in each of the worst four years fewer than half of
the estimated number of records were successfully geocoded.

One challenge with historical data is that geocoding may be impossible for some
records, for example if the address is a post office box, or if the street no
longer exists. However, there are several ways in which we could improve
geocoding success in future work.  A more robust named entity recognition
methodology could improve the ability to parse addresses from the text when
compared with the current approach relying on regular expressions. Also, we
could use additional features in Tesseract to increase the accuracy of the
characters returned during OCR. In particular, we could train Tesseract for the
different fonts which appear in the registries, and we could blacklist
characters that never appear in the registries.

\subsection{Application to Socio-environmental Studies}

This new approach to historical data collection has already made substantial
contributions to the sociological literature on urbanization and environmental
inequality \citep{elliott_historical_2013,lievanos_race_2015,mohai_which_2015};
and has direct relevance for industrial economics and geography as well
\citep{duranton_testing_2005,frenken_industrial_2015,martin_path_2006}. For
example, in their initial series of studies, Frickel and Elliott
(\citeyear{frickel_tracking_2008}; \citealp{elliott_environmental_2011,elliott_historical_2013,elliott_urbanization_2015})
also relied on state manufacturing registries, but practical concerns limited
data collection to the seven most polluting industrial sectors in the urban
core of their cities of interest. Combining this narrower subset of industrial
site data with tract-level data from the U.S. Census, as well as geocoded
hazardous site lists from state environmental regulatory agencies, made it
possible to conduct longitudinal analysis of demographic and regulatory
changes and how those dynamics influence the spatial and temporal accumulation
of environmental industrial risk.

\begin{figure*}[p]
\centering
\begin{tabulary}{\linewidth}{CC}
\includegraphics[width=\linewidth]{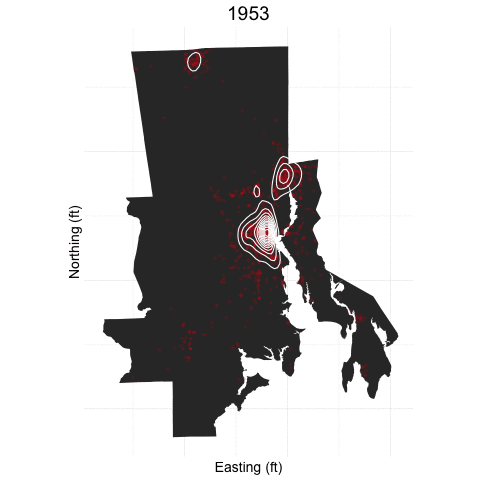} &
\includegraphics[width=\linewidth]{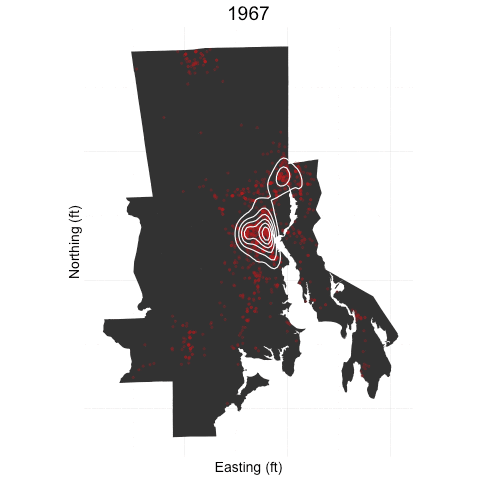} \\
\includegraphics[width=\linewidth]{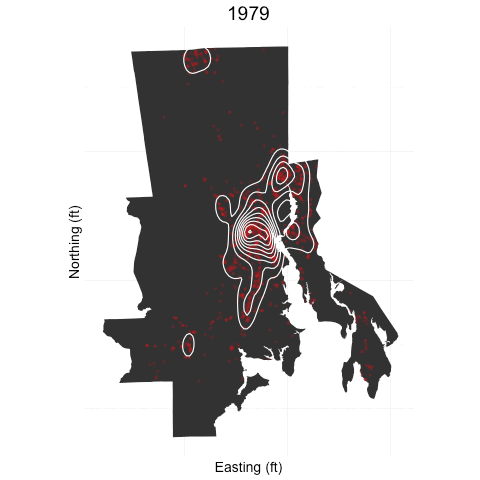} &
\includegraphics[width=\linewidth]{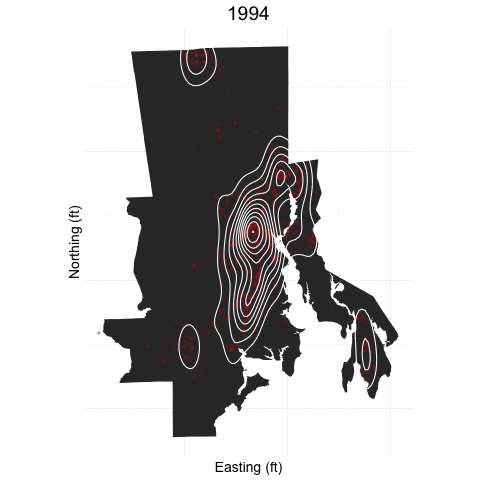}
\end{tabulary}
\caption{Density contour plots for all active manufacturing facilities.}
\label{fig:all}
\end{figure*}

\begin{figure*}[p]
\centering
\begin{tabulary}{\linewidth}{CC}
\includegraphics[width=\linewidth]{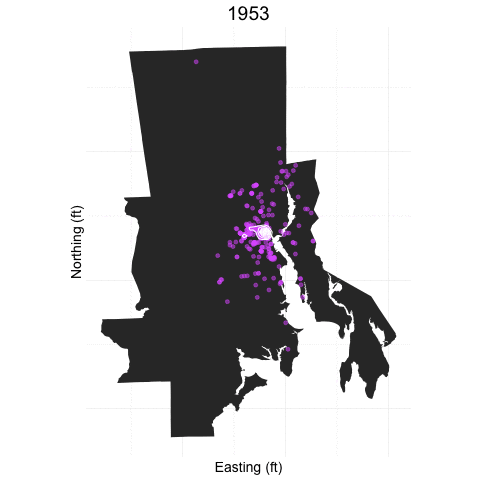} &
\includegraphics[width=\linewidth]{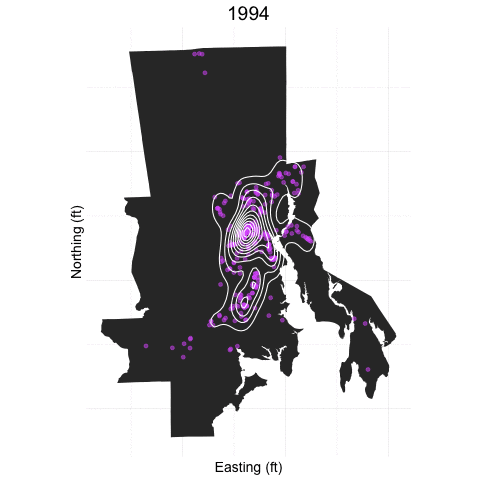} \\
\includegraphics[width=\linewidth]{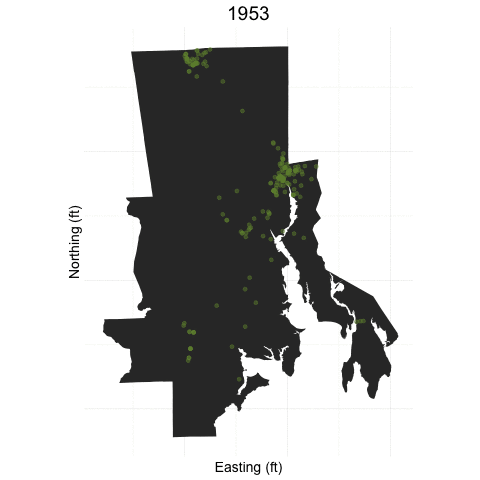} &
\includegraphics[width=\linewidth]{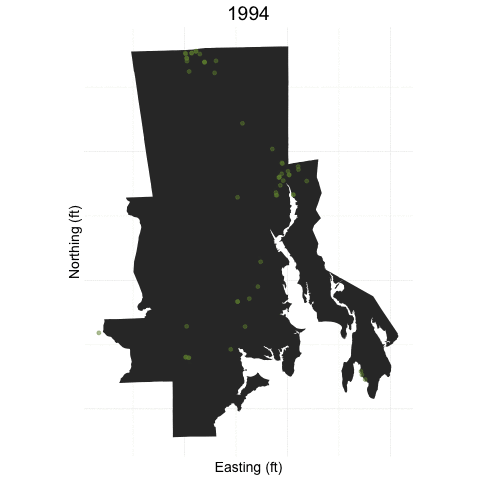}
\end{tabulary}
\caption{Density contour plots for active manufacturing facilities in the jewelry (purple) and textile (green) industries.}
\label{fig:jewelry}
\end{figure*}

By extending this previous approach to include all industrial sites in an
entire state over a 41 year period, we obtained several new results:

\begin{enumerate}

\item By including annual data for all manufacturing activities in Rhode
Island, we are able to study not only deindustrialization in the urban core of
Providence, but also the consequent suburbanization of manufacturing as
industrial activities gradually shifted to adjacent cities to the north and
south.

\item The additional spatial scope allows us to generate stronger
evidence that the spatial redistribution of manufacturing has been impacted by
changes in the transportation infrastructure of the state.
In 1953, manufacturing concentrated in the urban core of downtown Providence
near the large rail terminals and shipping port; by 1979, manufacturing had
begun to relocate southward following the newly constructed Interstate 95
corridor (Figure~\ref{fig:all}).

\item By mining data for all manufacturing sites, the new
approach has allowed greater flexibility in the exploration of sector-specific
clustering and distributional changes over time.
For example, preliminary analysis of the spatial and temporal distribution of
jewelry and textile manufacturing
(Figure~\ref{fig:jewelry}) -- two historically
important industries in Rhode Island -- shows that jewelry manufacturing
exhibits a distinct pattern of clustering while textiles have a more uniform
and widespread distribution.

\end{enumerate}

These preliminary findings illustrate
the advantages of our comprehensive approach to data mining, which
allows for richer empirical analysis of socio-environmental change and
greater opportunities for theory development than was possible in prior
research.

\section{Conclusion}

We have demonstrated a successful approach for mining and
geocoding structured text in historical registries, and shown its utility for
analyzing industrializion in the US state of Rhode Island.
Although we have applied these methods specifically to manufacturing
registries from Rhode Island, the same principles apply to other US state's
registries and to other printed historical documents with similar structure,
such as phone books, almanacs, and city directories. We are currently adapting
\texttt{georeg} to these additional data sources.
In particular, we believe phone books will be an ideal supplement,
adding commercial, non-profit (e.g. schools, hospitals, churches) and residential data to provide a comprehensive history of land use in urban areas.

\section*{Availability}
Source code for \texttt{georeg} is freely available for non-commercial use at \url{https://bitbucket.org/brown-data-science/georeg}.
All results presented in this paper were obtained using the version of \texttt{georeg} at commit hash \texttt{cacda11}.

\bibliographystyle{apa}
\bibliography{georeg}

\begin{thebibliography}{}

\bibitem[\protect\astroncite{Bradski}{2000}]{opencv_library}
Bradski, G. (2000).
\newblock {The OpenCV Library}.
\newblock {\em Dr. Dobb's Journal}, 25(11):120--126.

\bibitem[\protect\astroncite{Duranton and
  Overman}{2005}]{duranton_testing_2005}
Duranton, G. and Overman, H.~G. (2005).
\newblock Testing for {Localization} {Using} {Micro}-{Geographic} {Data}.
\newblock {\em The Review of Economic Studies}, 72(4):1077--1106.
\newblock doi:10.1111/0034-6527.00362\EatDot.

\bibitem[\protect\astroncite{Elliott and
  Frickel}{2011}]{elliott_environmental_2011}
Elliott, J.~R. and Frickel, S. (2011).
\newblock Environmental {Dimensions} of {Urban} {Change}: {Uncovering} {Relict}
  {Industrial} {Waste} {Sites} and {Subsequent} {Land} {Use} {Conversions} in
  {Portland} and {New} {Orleans}.
\newblock {\em Journal of Urban Affairs}, 33(1):61--82.
\newblock doi:10.1111/j.1467-9906.2010.00533.x\EatDot.

\bibitem[\protect\astroncite{Elliott and
  Frickel}{2013}]{elliott_historical_2013}
Elliott, J.~R. and Frickel, S. (2013).
\newblock The {Historical} {Nature} of {Cities} {A} {Study} of {Urbanization}
  and {Hazardous} {Waste} {Accumulation}.
\newblock {\em American Sociological Review}, 78(4):521--543.
\newblock doi:10.1177/0003122413493285\EatDot.

\bibitem[\protect\astroncite{Elliott and
  Frickel}{2015}]{elliott_urbanization_2015}
Elliott, J.~R. and Frickel, S. (2015).
\newblock Urbanization as {Socioenvironmental} {Succession}: {The} {Case} of
  {Hazardous} {Industrial} {Site} {Accumulation}.
\newblock {\em American Journal of Sociology}, 120(6):1736--1777.

\bibitem[\protect\astroncite{Frenken et~al.}{2015}]{frenken_industrial_2015}
Frenken, K., Cefis, E., and Stam, E. (2015).
\newblock Industrial {Dynamics} and {Clusters}: {A} {Survey}.
\newblock {\em Regional Studies}, 49(1):10--27.
\newblock doi:10.1080/00343404.2014.904505\EatDot.

\bibitem[\protect\astroncite{Frickel and Elliott}{2008}]{frickel_tracking_2008}
Frickel, S. and Elliott, J.~R. (2008).
\newblock Tracking {Industrial} {Land} {Use} {Conversions}: {A} {New}
  {Approach} for {Studying} {Relict} {Waste} and {Urban} {Development}.
\newblock {\em Organization \& Environment}, 21(2):128--147.
\newblock doi:10.1177/1086026608317799\EatDot.

\bibitem[\protect\astroncite{Liévanos}{2015}]{lievanos_race_2015}
Liévanos, R.~S. (2015).
\newblock Race, deprivation, and immigrant isolation: {The} spatial demography
  of air-toxic clusters in the continental {United} {States}.
\newblock {\em Social Science Research}, 54:50--67.
\newblock doi:10.1016/j.ssresearch.2015.06.014\EatDot.

\bibitem[\protect\astroncite{Lloyd}{1982}]{lloyd_least_1982}
Lloyd, S.~P. (1982).
\newblock {Least squares quantization in PCM}.
\newblock {\em IEEE Transactions on Information Theory}, 28:129--137.
\newblock doi:10.1109/TIT.1982.1056489\EatDot.

\bibitem[\protect\astroncite{Ma and Doermann}{2003}]{Ma_Doermann_2003}
Ma, H. and Doermann, D.~S. (2003).
\newblock Bootstrapping structured page segmentation.
\newblock {\em Proc. SPIE}, 5010:179--188.
\newblock doi:10.1117/12.476058\EatDot.

\bibitem[\protect\astroncite{Mao et~al.}{2003}]{Mao_Rosenfeld_Kanungo_2003}
Mao, S., Rosenfeld, A., and Kanungo, T. (2003).
\newblock Document structure analysis algorithms: a literature survey.
\newblock {\em Proc. SPIE}, 5010:197--207.
\newblock doi:10.1117/12.476326\EatDot.

\bibitem[\protect\astroncite{Martin and Sunley}{2006}]{martin_path_2006}
Martin, R. and Sunley, P. (2006).
\newblock Path dependence and regional economic evolution.
\newblock {\em Journal of Economic Geography}, 6(4):395--437.
\newblock doi:10.1093/jeg/lbl012\EatDot.

\bibitem[\protect\astroncite{Mohai and Saha}{2015}]{mohai_which_2015}
Mohai, P. and Saha, R. (2015).
\newblock Which came first, people or pollution? {Assessing} the disparate
  siting and post-siting demographic change hypotheses of environmental
  injustice.
\newblock {\em Environmental Research Letters}, 10:115008.
\newblock doi:10.1088/1748-9326/10/11/115008\EatDot.

\bibitem[\protect\astroncite{Smith}{2007}]{Smith_2007}
Smith, R.~W. (2007).
\newblock {An Overview of the Tesseract OCR Engine}.
\newblock In {\em {12th International Conference on Document Analysis and
  Recognition}}, pages 629--633, Los Alamitos, CA, USA.
\newblock doi:10.1109/ICDAR.2007.56\EatDot.

\bibitem[\protect\astroncite{Smith}{2009}]{Smith_2009}
Smith, R.~W. (2009).
\newblock {Hybrid Page Layout Analysis via Tab-Stop Detection}.
\newblock In {\em {10th International Conference on Document Analysis and
  Recognition}}, pages 241--245, Barcelona, Catalonia, Spain.
\newblock doi:10.1109/ICDAR.2009.257\EatDot.

\bibitem[\protect\astroncite{Suzuki and Abe}{1985}]{suzuki_topological_1985}
Suzuki, S. and Abe, K. (1985).
\newblock Topological structural analysis of digitized binary images by border
  following.
\newblock {\em Computer Vision, Graphics, and Image Processing}, 30(1):32--46.
\newblock doi:10.1016/0734-189X(85)90016-7\EatDot.

\bibitem[\protect\astroncite{Tang et~al.}{1996}]{Tang_Lee_Suen_1996}
Tang, Y.~Y., Lee, S.-W., and Suen, C.~Y. (1996).
\newblock {Automatic document processing: A survey}.
\newblock {\em Pattern Recognition}, 29(12):1931--1952.
\newblock doi:10.1016/S0031-3203(96)00044-1\EatDot.

\end{thebibliography}

\end{document}